\documentclass[10pt]{article}

\usepackage[utf8]{inputenc}
\usepackage[T1]{fontenc}

\title{\LARGE \bf
	A Preliminary Study for a Quantum-like Robot Perception Model%
	\thanks{Paper submitted to \textit{29th IEEE International Conference on Robot \& Human Interactive Communication (RO-MAN)}, 2020, Naples, Italy.}
}

\author{%
	Davide Lanza%
	\thanks{%
		Department of Informatics, Bioengineering, Robotics, and Systems Engineering, University of Genoa, Via All'Opera Pia 13, 16145, Genoa, Italy.
		{\tt\footnotesize davide.lanza@eleves.ec-nantes.fr},
		{\tt\footnotesize fulvio.mastrogiovanni@unige.it}}
	\and Paolo Solinas%
	\thanks{%
		Department of Physics, University of Genoa, and National Institute for Nuclear Physics (Genoa section), Via Dodecaneso 33, 16146, Genoa, Italy.
		{\tt\footnotesize solinas@fisica.unige.it}}
	\and Fulvio Mastrogiovanni\footnotemark[2]
}

\date{}

\usepackage{graphicx, pgf, float, caption, subcaption} 
\usepackage{amsmath, amsfonts, bm, mathtools, cancel} 
\usepackage{color, xcolor} 
\usepackage{multirow, longtable, numprint} 

\usepackage{hyperref}
\hypersetup{%
	pdftoolbar=true,        
	pdfmenubar=true,        
	pdffitwindow=false,     
	pdfstartview={FitH},    
	pdftitle={A Preliminary Study for a Quantum-like Robot Perception Model},    
	pdfauthor={Davide Lanza},     
	pdfsubject={Subject},   
	pdfcreator={Davide Lanza},   
	pdfproducer={Davide Lanza}, 
	pdfkeywords={Robotics} {Quantum} {Perception} {Modeling}, 
	pdfnewwindow=true,      
	colorlinks=true,       
	linkcolor=blue,          
	citecolor=blue,        
	filecolor=blue,      
	urlcolor=blue,           
}

\newcommand{\ket}[1]{
	\left| #1 \right\rangle
}
\newcommand{\plotscalingfactor}{0.23}

\begin{document}

\maketitle
\pagestyle{headings}

\begin{abstract}

Formalisms based on quantum theory have been used in Cognitive Science for decades due to their descriptive features. 
A quantum-like (QL) approach provides descriptive features such as state superposition and probabilistic interference behavior. 
Moreover, quantum systems dynamics have been found isomorphic to cognitive or biological systems dynamics. 

The objective of this paper is to study the feasibility of a QL perception model for a robot with limited sensing capabilities. 
We introduce a case study, we highlight its limitations, and we investigate and analyze actual robot behaviors through simulations, while actual implementations based on quantum devices encounter errors for unbalanced situations. 
In order to investigate QL models for robot behavior, and to study the advantages leveraged by QL approaches for robot knowledge representation and processing, we argue that it is preferable to proceed with simulation-oriented techniques rather than actual realizations on quantum backends.
\end{abstract}


\section{Introduction}
\label{sec:intro}

In the last decades, the quantum mechanics formalism has been studied and applied outside its initial scope. 
Regarding human-level perception and cognition modeling, it has been used to address biological problems at diverse macroscopic scales. 
Quantum-like (QL) probability models have been analyzed and applied to Biology as well as Cognitive Science and decision theory. 
High-level applications involving cognition aspects and decision making modeling include 
bistable perceptions \cite{manousakis_quantum_2009}, 
non-compositional concept representation \cite{aerts_theory_2005, blutner_concepts_2009},
human-like information processing \cite{busemeyer_quantum_2012},
conjoint memory recognition \cite{busemeyer_quantum_2012}, 
human semantic space \cite{aerts_quantum_2004},
and
quantum learning \cite{ivancevic_quantum_2010}.

In this paper, we describe a low-level limited perception model that employs a QL approach to store perceptual information. 
The main contribution of this study is 
(i) the designed QL perception model, subject of course to a series of modelling assumptions and limitations, and 
(ii) an assessment of current possibilities leading to its physical implementation on existing quantum computers. 
The experimental results reported here seem to confirm the feasibility of our approach. 
We obtained the designed behavior through simulations on a classical device, while actual quantum computer implementations registered notable errors. 
In our opinion, then, at the current stage of quantum computers development and engineering, it is preferable to proceed in further studies with simulation-oriented techniques rather than actual realizations on quantum backends, because of the lack of optimization of current quantum computing hardware for such unprecedented applications. 

In Section \ref{sec:stateofart} are reviewed the main achievements of the QL approach to perception and cognition modeling. 
Section \ref{sec:modeldef} presents the proposed QL model for robot perception, while in Section \ref{sec:expsetting} the experimental results are reported.
These results are not to be considered as definitive findings on the accuracy of the model, but a preliminary feasibility check, as discussed in Section \ref{sec:discussion}.


\section{Quantum-like Perception Modeling}
\label{sec:stateofart}

Since the early intuitions by Amann \cite{amann_gestalt_1993}, much work has been carried out regarding the perception of impossible figures and the isomorphism between consciousness and quantum dynamics in these cases. 
Conte \cite{conte_testing_2008} studied extensively quantum interference effects in human perception and cognition of ambiguous figures, concluding that mental states are compatible with Quantum Mechanics \cite{conte_mental_2009}. 
Manousakis \cite{manousakis_quantum_2009} used a quantum-inspired formalism to mathematically describe simple perception processes, and in particular to describe the probability distribution of perceptive dominances in subjects experiencing binocular rivalry. 
In this model, the two alternating perceptions were associated with the basis states of a two-state quantum system. 
While this model has been criticized for not taking a mixed-perception state \cite{paraan_more_2014} into account, our model described in Section \ref{sec:modeldef} is inspired by Manousakis' perception mapping to basis states. 



Caves \textit{et al.} \cite{caves_quantum_2002} proposed the interpretation of quantum probability theory with a Bayesian approach, i.e., probabilities quantify a degree of belief for a single trial, with no \textit{a priori} connection to limiting frequencies. 
In that view, the classical quantum distinction for probabilities relies on the nature of the information they encode rather than on their definition. 
In fact, while having the maximal amount of information about a certain system in a classical world means having complete knowledge about its behavior, in a quantum \emph{world} the maximal amount of information cannot be complete. 
The model we developed encodes perceptual information in a single qubit. 
Following this interpretation, the information carried by the qubit represents the degree of belief of the robot for a single trial, which is measuring the qubit itself.

While these are the contributions to the literature that inspired our model the most, many applications have been studied over the years and unfold new perspectives for higher complexity models in further research. 
QL approaches have been applied to semantic analysis \cite{bruza_quantum_2005}, 
human information processing \cite{busemeyer_quantum_2012}, 
and the human semantic space \cite{aerts_quantum_2004}.
All these works can inspire multi-perceptual integration in higher-level models. 
Moreover, human motivation has been modeled as well \cite{ivancevic_life-space_2007}, a quantum Belief-Desire-Intention (BDI) model for information processing and decision-making has been developed \cite{bisconti_quantum-bdi_2015}, and QL decision-making models have been used to study human judgment probabilities \cite{bordley_quantum_1998, bordley_experiment-dependent_1999}.


\section{Model Definition}
\label{sec:modeldef}

\begin{figure}[t]
	\centering
	\includegraphics[height=3cm]{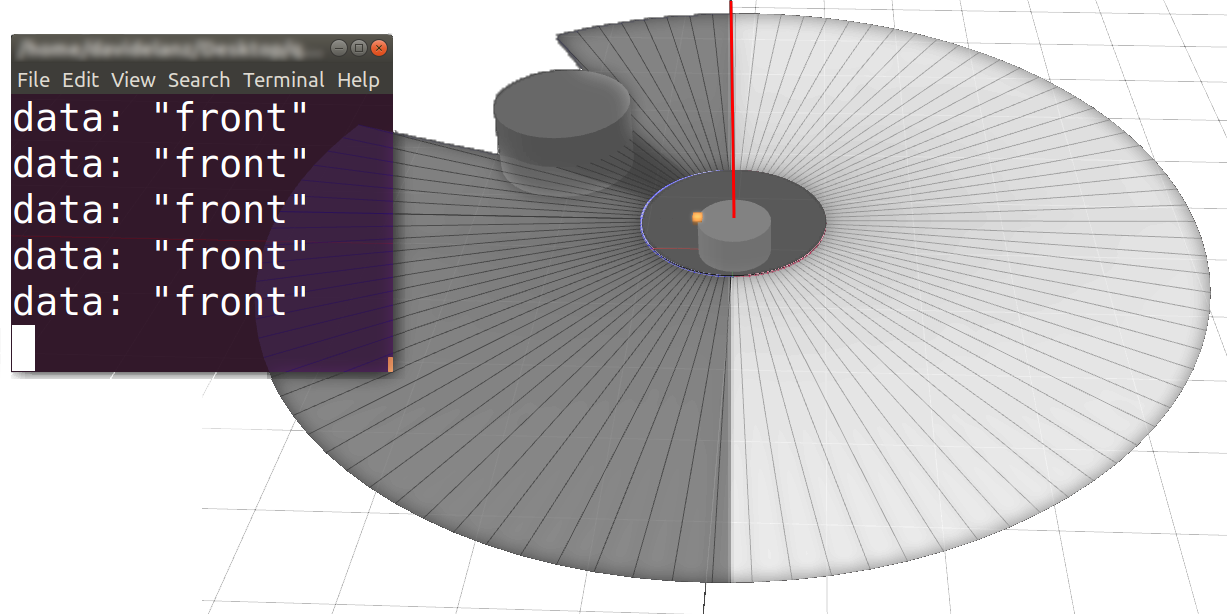}
	\hspace{20pt}
	\includegraphics[height=3cm]{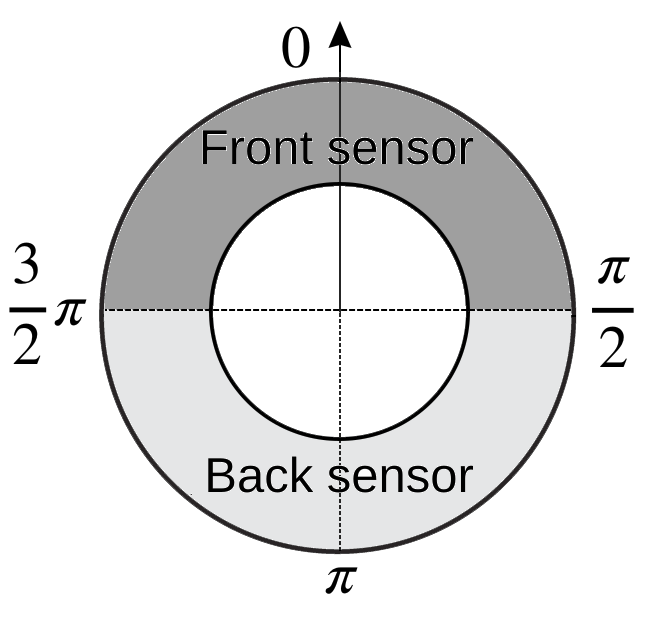}
	\caption{A case study in simulation.}
	\label{fig:2statesgazebo}
\end{figure}

\subsection{A Case Study}
\label{subsec:case-study}

For our model, we consider a circle-shaped robot positioned in a flat environment, containing one cylindrical object randomly moving along a plane (Fig. \ref{fig:2statesgazebo}). 
The robot's position and orientation are fixed, i.e., the robot does not perform any movement. 
The object moves in a way such that it never collides with the robot. 
The robot is equipped with two presence sensors are able to detect whether something is in its measurement range. 
The ``front'' sensor's span is $(-\frac{\pi}{2},\frac{\pi}{2})$, while the ``back'' sensor's span is $(\frac{\pi}{2},\frac{3}{2}\pi)$ (Fig. \ref{fig:2statesgazebo}). 
The object moves in such a way that it never goes too far from being detected by one of the two sensors.
As a consequence, for each measurement the robot always acquires only one measurement, namely a single ``front'' event or a single ``back'' event. 
Each event is represented with the corresponding angle $\alpha$, i.e., $\alpha = 0$ for a ``front'' event, and $\alpha = \pi$ for a ``back'' event. 

The robot acquires sensory data with a sample period of $T_s$ seconds. 
For this study, we consider $\tau$ events in a $\left.\Delta t = \tau T_s\right.$ time span. 
Therefore, we consider an ordered sequence $\Sigma$ of successive events $\alpha_{k\in[1,\tau]}$ as the one in the following figure:
\begin{center}
	\includegraphics[width=.65\linewidth]{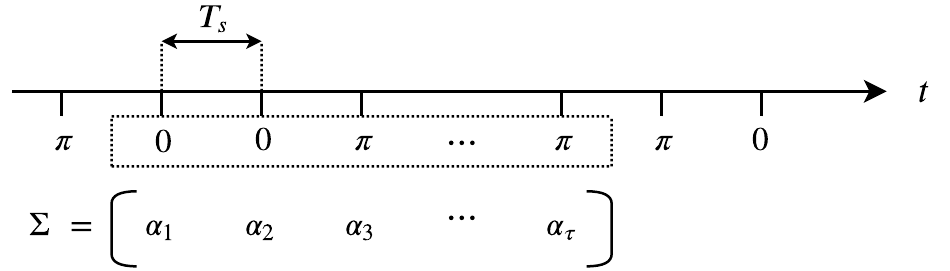}
\end{center}
Given a function $count_{x \in A}(x=y)$ that returns the number of elements in a discrete set $A$ which are equal to $y$, we define the relative frequency of an event $\bar{\alpha}$ as
\begin{equation}
	f(\Sigma, \bar{\alpha})
	=
	\frac{
		\underset{\alpha_k \in \Sigma}{count}(\alpha_k = \bar{\alpha})
	}
	{
		\tau
	}.
	\label{eq:relfreq}
\end{equation}
The relative frequency $f$ of an event is the information we are interested in to encode in the qubit, representing -- as anticipated in Section \ref{sec:stateofart} -- the degree of belief of the robot for a certain event \cite{caves_quantum_2002}. 
For the case study presented here, given $\Sigma$ we only need to consider $f$ for one event, for example $f(\Sigma, \pi)$ since $f(\Sigma, 0) = 1-f(\Sigma, \pi)$.

According to the limitations posed above, we have two world states defined by the robot sensory capabilities, i.e., the moving object is in front of the robot or behind the robot. 
Such possible states are related to the robot's perceived world projection rather than the simulated environment, hence we can call these states ``robot's perception states''. 
In fact, we are not considering the relative position of the moving object with respect to the robot, but just its relative collocation with respect to the robot's sensory data. 
As Manousakis \cite{manousakis_quantum_2009} does for binocular rivalry perception models, we map the two mutually exclusive perception states to the two base states of a single qubit.


\subsection{Information Encoding}
\label{subsec:info-enc}

A qubit is a mathematical object which, like a classical bit, has a state. 
While classical bit states can assume only two values ($0$ and $1$), a qubit state $\ket{\psi}$ is a linear combination, called ``superposition'', of its two basis states
\begin{equation}
	\ket{\psi}
	=
	c_0
	\ket{0}
	+
	c_1
	\ket{1},
	\label{eq:lincomb}
\end{equation}
where $c_0$ and $c_1$ are complex numbers. 
Unlike a classical bit, the qubit state cannot be accessed.
When a measurement is performed on $\ket{\psi}$, the result is either $0$ with probability $|c_0|^2$ or $1$ with probability $|c_1|^2$. 
As a consequence it follows that 
\begin{equation}
	|c_0|^2 + |c_1|^2 = 1.
	\label{eq:bornpostulate}
\end{equation}
After each measurement, $\ket{\psi}$ collapses on the measured basis state, and all the information about $c_1$ and $c_2$ is lost \cite{nielsen_quantum_2010}. 
This is why the interpretation by Caves \textit{et al.} \cite{caves_quantum_2002} of the information carried by the qubit is akin to the degree of belief of the robot for a single trial, i.e., the measurement on the qubit itself (Section \ref{sec:stateofart}). 
Since there is no direct correspondence in Quantum Mechanics between a quantum system's state and the measurement performed on it, it is impossible to predict a single measurement output. 
However, information carried by qubits can be manipulated to obtain measurements which depend distinctly on the properties of the state \cite{nielsen_quantum_2010}. 
For this preliminary study, we have not investigated these indirect methods. 
Instead, in order to obtain information about $|c_0|^2$ and $|c_1|^2$, we performed $N$ measurements on $N$ identically prepared qubits. 
If we define $\widehat{|c_i|^2}$ as the relative frequency of measuring $\ket{i}$, since an event's probability can be interpreted as the limit of its relative frequency \cite{friedman_frequency_1999} we have
\begin{equation}
	\lim\limits_{N\to\infty}
	\widehat{|c_i|^2}
	=
	|c_i|^2,
	\quad
	i \in \{0,1\}.
	\label{eq:frequency-limit}
\end{equation}

A single qubit state can be easily represented geometrically in a Bloch sphere. 
In fact \eqref{eq:lincomb} can be rewritten \cite{nielsen_quantum_2010} as
\begin{equation}
	\ket{\psi}
	=
	\cos 
	\left(\frac{\theta}{2}\right)
	\ket{0}
	+
	e^{i\varphi }\sin 
	\left(\frac{\theta}{2}\right)
	\ket{1},
	\label{eq:bloch_coords}
\end{equation}
where $\theta$ and $\varphi$ are real numbers. 
In this representation, the qubit state $\ket{\psi}$ is a unitary vector that points one of the points on the sphere's surface, as illustrated in Fig. \ref{fig:blochsphere}.

As mentioned in Section \ref{subsec:case-study}, following the approach that Manousakis \cite{manousakis_quantum_2009} uses to define the binocular rivalry perception model, we mapped the two, mutually exclusive perception states to the two base states of a single qubit. 
Hence, the ``front'' state corresponds to the basis state $\ket{0}$, and the ``back'' state to $\ket{1}$. 
We want to encode the relative frequency $f$ of the two events in the qubit. 
First, we define the number of events associated with the basis state $\ket{i_{\in\{0,1\}}}$ as
\begin{equation}
	\tau_{\ket{0}} = \underset{\alpha_k \in \Sigma}{count}(\alpha_k = 0)
	,\;\;
	\tau_{\ket{1}} = \underset{\alpha_k \in \Sigma}{count}(\alpha_k = \pi).
\end{equation}
This allows us to redefine the relative frequency of an event \eqref{eq:relfreq} in a more convenient way\footnote{
	In the following equations, as for $\tau$ in \eqref{eq:relfreq}, the values $f_{\ket{i}}$ and $\tau_{\ket{i}}$ depend on a particular sequence $\Sigma$, but the dependency has been omitted to allow for a lighter notation.
}:
\begin{equation}
	f(\Sigma, 0)
	:=
	f_{\ket{0}}
	=
	\frac{\tau_{\ket{0}}}{\tau}
	, \quad
	f(\Sigma, 1)
	:=
	f_{\ket{1}}
	=
	\frac{\tau_{\ket{1}}}{\tau}
	.
\end{equation}
Since $f_{\ket{0}}= 1 - f_{\ket{1}}$, we encode just the $f_{\ket{1}}$ in the qubit as
\begin{equation}
	\theta = \pi f_{\ket{1}},
	\label{eq:encoding}
\end{equation}
where $\theta$ refers to the Bloch sphere representation of $\ket{\psi}$, and an example is shown in Fig. \ref{fig:demosequence}. 
To do so, we initialize the qubit $\ket{\psi}$ at $\ket{0}$ and we use a unitary operator $U$ to apply a fractional rotation of $\pi/\tau$ along the $y$ axis of the Bloch sphere representation. 
This operator has to be applied to the qubit $\tau_{\ket{1}}$ times for a sequence of events $\Sigma$, in order to obtain the desired encoding \eqref{eq:encoding}\footnote{A demonstration video is available at \href{https://www.youtube.com/watch?v=EvE24PCdU8E}{https://youtu.be/EvE24PCdU8E}.}.

\subsection{Operator Definition}

Since operations on a qubit state must preserve its norm, they have to be described by $2\times2$ unitary matrices \cite{nielsen_quantum_2010}. 
Unitary operators are used in quantum theory to formalize the evolution of a system, that is, in this case, the qubit $\ket{\psi}$. 
Given a normalized \eqref{eq:bornpostulate} state vector $\ket{\psi}$ encoding a certain probability distribution, it can be pre-multiplied by a unitary operator $U$ producing a new state vector $U \ket{\psi}$ which is still normalized and encodes a new probability distribution. 
Unitary operators are reversible and can be chain-multiplied together to represent a sequence of evolution steps \cite{nielsen_quantum_2010}, such as
\begin{equation}
	\ket{\psi}' 
	=
	\prod_{k=1}^{\tau}U_k \ket{\psi}
	=
	U_{tot} \ket{\psi},
\end{equation}
and for this reason we use them to encode the event relative frequency in $\ket{\psi}$. 
Pauli matrices are a useful set of matrices in quantum mechanics, written in the $\{\ket{0},\ket{1}\}$ basis as
\begin{equation}
\sigma_x
=
\begin{bmatrix} 0 & 1 \\ 1 & 0 \end{bmatrix},
\;
\sigma_y
=
\begin{bmatrix} 0 & -i \\ i & 0 \end{bmatrix},
\;
\sigma_z
=
\begin{bmatrix} 1 & 0 \\ 0 & -1 \end{bmatrix}.
\end{equation}
When they are exponentiated, Pauli matrices generate three important classes of unitary matrices, which are the rotation operators about the three axes \cite{nielsen_quantum_2010}. 
This is possible because rotations belong to the special unitary group $SU(2)$, that is the Lie group of $2 \times 2$ unitary matrices with unitary determinant \cite{hall_lie_2015}, i.e.,
\begin{equation}
	SU(2)=
	\left\{
	\begin{bmatrix}
		a & -{\overline {b}} \\
		b & {\overline {a}}
	\end{bmatrix}
	\
	a, b \in \mathbb {C},
	|a|^{2}+|b|^{2}=1
	\right\}.
\end{equation}
The Lie algebra of a group is the set of all matrices $A$ such that $\exp(x A)$ is an element of that group for all real numbers $x$. 
The Lie algebra $\mathfrak{su}(2)$ of $SU(2)$ consists of $2 \times 2$ skew-Hermitian matrices with zero trace \cite{hall_lie_2015}, i.e.,
\begin{equation}
	\mathfrak {su}(2)
	=
	\left\{
		\begin{bmatrix}
			i a & -{\overline{b}}\\
			b&-i\ a
		\end{bmatrix}
		\ a\in \mathbb {R} ,b\in \mathbb {C} 
	\right\},
\end{equation}
whereas Pauli matrices multiplied by $i$ form a basis for it, such as
\begin{equation}
	\{
	i\sigma_x,
	i\sigma_y,
	i\sigma_z
	\}.
\end{equation}
As mentioned above, the elements of this set generate rotation operators when exponentiated.
We are interested in the rotation operator around the $y$ axis, defined as
\begin{equation}
R_y(\theta) = \exp\left({-i\frac{\theta\sigma_y}{2}}\right).
\label{eq:rotation1}
\end{equation}
Since $\sigma_y$ is a unitary matrix, it is possible to rewrite \eqref{eq:rotation1} according to the Euler rotation theorem, such as
\begin{equation}
\begin{split}
R_y(\theta)
=
\cos \frac{\theta}{2} \mathbb{I} 
- i \sin \frac{\theta}{2} \sigma_y
=
\begin{bmatrix}
\cos \frac{\theta}{2}
& 
-\sin \frac{\theta}{2}
\\
\sin \frac{\theta}{2}
& 
\cos \frac{\theta}{2}
\end{bmatrix}.
\end{split}
\end{equation}

\begin{figure}[t!]
	\centering
	\begin{minipage}{.45\linewidth}
		\centering
		\includegraphics[height=3.2cm]{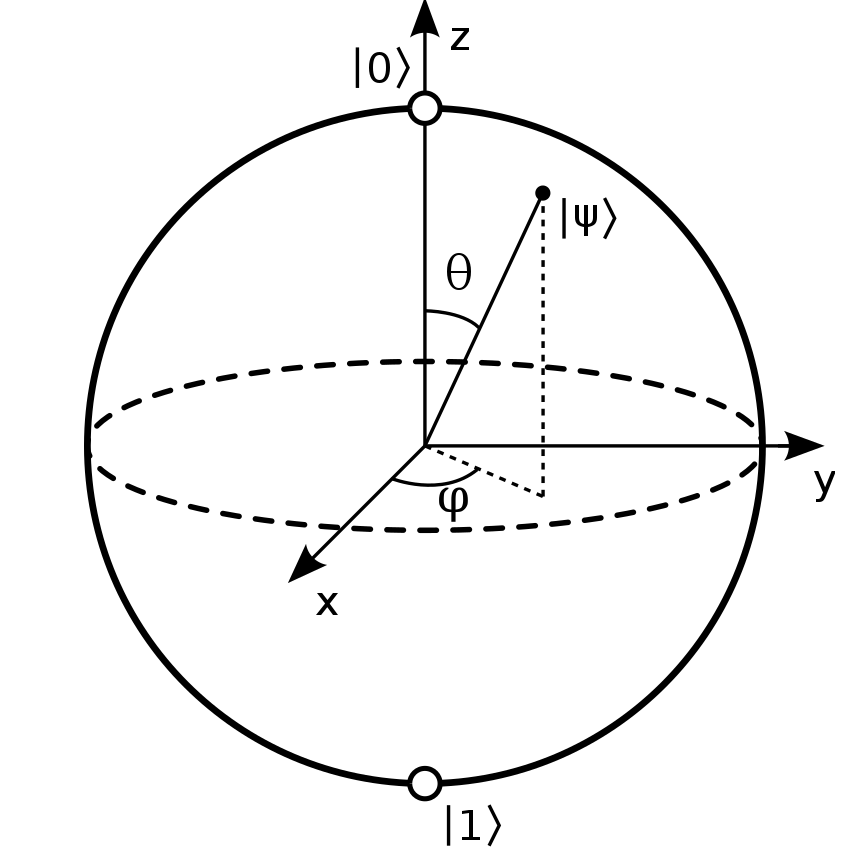}
		\captionof{figure}{Bloch sphere.}
		\label{fig:blochsphere}
	\end{minipage}
	\begin{minipage}{.45\linewidth}
		\centering
		\includegraphics[height=3.2cm]{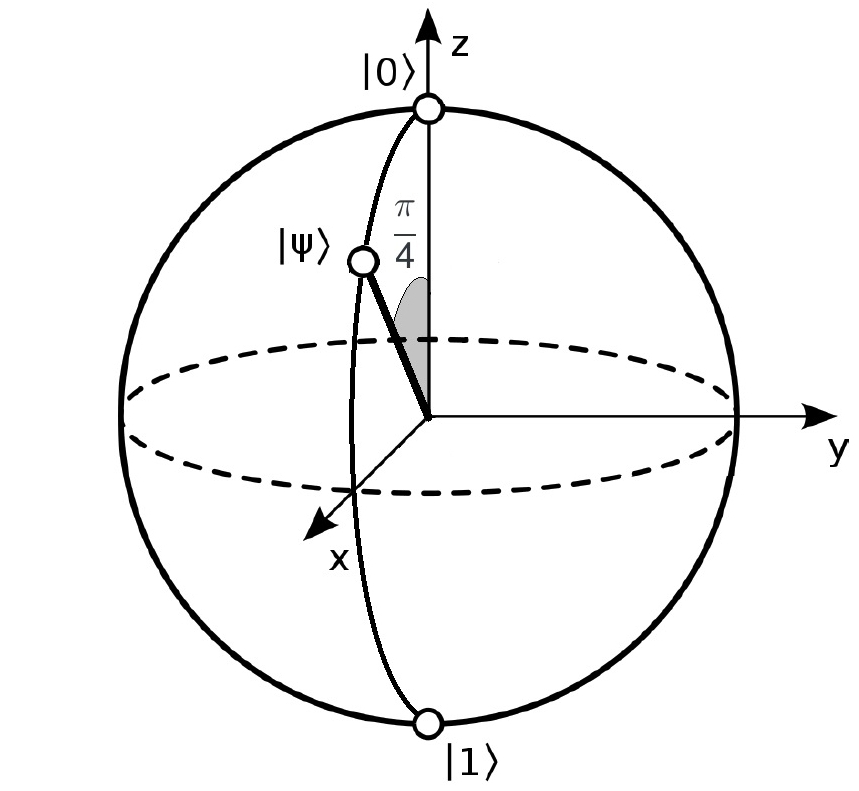}
		\captionof{figure}{$\ket{\psi}$ for $f_{\ket{1}}=\frac{1}{4}$.}
		\label{fig:demosequence}
	\end{minipage}%
\end{figure}

The operator $U$ applies $R_y(\frac{\pi}{\tau})$ to $\ket{\psi}$ for each registered event $\left.\alpha = \pi\right.$, while for every event $\alpha = 0$ does not. Hence, given $\alpha_k \in \Sigma$, operator $U$ can be defined as
\begin{align}
	U_k
	&
	=
	\begin{cases}
		\mathbb{I}
		&
		\text{ if } \alpha_k = 0,  
		\\
		\exp \Big(
			-i\cfrac{\pi \sigma_y}{2\tau}
		\Big)
		&
		\text{ if } \alpha_k = \pi,
		\\
	\end{cases}
	\label{eq:operator}
\end{align}
which, for this particular mapping, can be simplified as
\begin{equation}
	U_k
	=
	\exp\left(
	-i\frac{\alpha_k \sigma_y}{2\tau}
	\right).
	\label{eq:operator_short}
\end{equation}
For example, given a sequence $\Sigma$ such that $\tau=12$ and $\tau_{\ket{1}} = 3$ we obtain
\begin{equation}
U_{tot}
=
\prod_{k=1}^{\tau}
U_k
=
\exp\left(
-i\ \frac{\ 3\ \sigma_y}{24\ \tau}
\right)
=
R_y\left(
\frac{\pi}{4}
\right),
\end{equation}
and the corresponding qubit state is illustrated in Fig. \ref{fig:demosequence}.

It is noteworthy that this definition relies on the assumption that the initial state is the ``back'' state $\ket{0}$. 
If preferred, one can initialize the qubit in a balanced superposition state with an Hadamard gate \cite{nielsen_quantum_2010} to have as initial state
\begin{equation}
	\ket{+}
	=
	\frac{1}{\sqrt{2}}\ket{0}
	+
	\frac{1}{\sqrt{2}}\ket{1}
	.
\end{equation}
In this case, one should then use:
{\small
\begin{align}
	U'
	&
	=
	\begin{cases}
	\exp \Big(
	+i\cfrac{\pi \sigma_y}{4\tau}
	\Big)
	&
	\text{ if } \alpha = 0,  
	\\
	\exp \Big(
	-i\cfrac{\pi \sigma_y}{4\tau}
	\Big)
	&
	\text{ if } \alpha = \pi. 
	\\
	\end{cases}
\end{align}
}

\subsection{Information Decoding for Validation Purposes}

As we pointed out in Section \ref{subsec:case-study}, there exist indirect techniques useful to exploit the information which $\ket{\psi}$ encodes to some extent \cite{nielsen_quantum_2010}\cite{ruppert_martingale_2010}\cite{hangos_state_2011}. 
For this preliminary study, we decided not to investigate these indirect methods. 
Instead, in order to obtain information about $|c_0|^2$ and $|c_1|^2$, we performed $N$ measurements on $N$ identically prepared experiments. 
As shown in \eqref{eq:frequency-limit}, for $N$ big enough, the measured relative frequencies $\widehat{|c_i|^2}$ (with $i\in\{0,1\}$) \emph{tend} to the expected values $|c_i|^2$.

Since we are interested in the relative frequency $f_{\ket{1}}$, in order to compute the approximation error we have to convert the measured values. 
Recalling \eqref{eq:lincomb}, \eqref{eq:bloch_coords}, and \eqref{eq:encoding}, we have
\begin{equation}
|c_1|^2
=
\sin^2 
\left(
\frac{\pi}{2}f_{\ket{1}}
\right),
\end{equation}
because $\varphi=0$ for all possible $\Sigma$, since the rotations are performed only around $y$. 
We can then define the empirical relative frequency as
\begin{equation}
	\widehat{f}_{\ket{1}}
	=
	\frac{2}{\pi}
	\arcsin
	\left(
	\sqrt{\widehat{|c_1|^2}}
	\right).
	\label{eq:correction}
\end{equation}
We define the decoding error as
\begin{equation}
\varepsilon
	=
	|\;
	f_{\ket{1}}
	-
	\widehat{f}_{\ket{1}}
	\;|
	\label{eq:error}
\end{equation}
Hence, $\widehat{f}_{\ket{1}}$ is the information we obtain from the decoding of $\ket{\psi}$. 
This decoding approach is useful only for offline validation purposes, since $N$ identical experiments are necessary every time. 
In an online scenario, this would mean that the operator $U$ should be applied $N\tau$ times, leading to $N\tau_{\ket{1}}$ fractional rotations for each $\Sigma$ (collected every $\Delta t$). 
This is highly unpractical since we need high values of $N$ in order to get small $\varepsilon$ values. 
Moreover, if we use only one qubit, we should re-initialize it to $\ket{0}$ for each of the $N$ iterations, or we would need $N$ qubits identically processed in parallel.


However, since the goal of this preliminary study is not an online implementation of an actual quantum model, we decided to adopt this decoding approach in order to study the feasibility of the model, rather than implementing it in an online setting. 
This choice is motivated by its simplicity, since this $N$ measurement state estimation process is a basic functionality exhibited by the majority of currently available quantum frameworks, while indirect measurement techniques need dedicated research.


\section{Model Simulation}
\label{sec:expsetting}

\begin{figure}
	\centering
	\includegraphics[height=2.2cm]{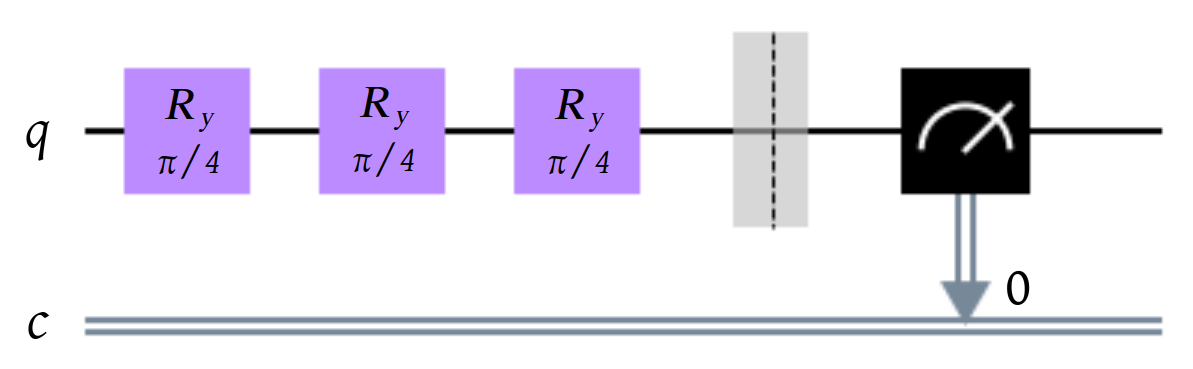}
	\caption{Circuit for $\tau=4, \tau_{\ket{1}}=\;3$}
	\label{fig:democirc}
\end{figure}

\subsection{From Sequences to Circuits}

In order to test the feasibility of our model with respect to the target case study, we simulated the robot behavior estimating different $\widehat{f}_{\ket{1}}$ for different sequences $\Sigma$, comparing the corresponding approximation errors $\varepsilon$. 
For the implementation, we rely on the IBM Quantum Experience (IBMQ) environment \cite{noauthor_ibm_nodate}. 
IBMQ provides Qiskit, i.e., an open-source quantum computing software development framework \cite{noauthor_ibm_nodate-1}. 
Qiskit's workflow consists of three high-level steps. 
First, designing the quantum circuit that represents the problem. 
Then, run experiments on different backends ($N$ iterations approach) including local simulators and cloud-based quantum computer backends. 
Finally, analyzing the data collected from the executed runs.

Considering a sequence $\Sigma$ and the corresponding $\tau$ and $\tau_{\ket{1}}$ values, the circuit can be designed applying $\tau_{\ket{1}}$ operation $R_y(\pi/\tau)$ on the qubit $q$ initialized to $\ket{0}$, as shown for the example in Fig. \ref{fig:democirc}. 
In order to run the experiments, a measurement gate is applied on the qubit $q$, mapping the output on the classical bit $c$ (Fig. \ref{fig:democirc}). 
Then, it is possible to process the circuit through a IBMQ backend, which runs $N$ experiments, returning $N_0$ for each experiment resulted in a classical bit outcome $c=0$, and $N_1$ for each one resulted in a $c=1$. 
Hence, recalling \eqref{eq:correction} we can write
\begin{equation}
	\widehat{|c_i|^2} = \frac{N_i}{N}
	,\quad
	\widehat{f}_{\ket{i}}
	=
	\frac{2}{\pi}
	\arcsin
	\left(
	\sqrt{\frac{N_i}{N}}
	\right).
\end{equation}

\subsection{IBMQ Backends}

\begin{table}
	\centering
	\caption{
		IBMQ backend calibrations (Apr 19, 2020)
	}
	\npdecimalsign{.}
	\nprounddigits{3}
	\tabcolsep2mm 
	\renewcommand{\arraystretch}{1.1}
	\resizebox{.7\linewidth}{!}{%
		\begin{tabular}{ c 
				N{3}{3} 
				N{3}{3} 
				N{3}{1} 
				N{3}{1}
				N{2}{3}}
			\hline\noalign{\smallskip}
			Qubit&
			\text{$T_1$ ($\mu s$)} &
			\text{$T_2$ ($\mu s$)} &
			\parbox{1.4cm}{
				\centering
				Frequency\\(GHz)} &
			\parbox{1.4cm}{
				\centering
				Readout error
			} &
			\parbox{2.2cm}{
				\centering
				Single-qubit U2 error rate
			}\\[6pt]
			\hline
			\multicolumn{6}{l}{\textbf{Armonk}:}\\
			$q_0$ &
			144.311046943353 &
			138.107119092533 &
			4.97429712381706 &
			0.0815 &
			0.790975078538e-3\\
			\hline
			\multicolumn{6}{l}{\textbf{Burlington}:}\\
			$q_0$ &	
			96.9447645597802&
			48.3591245265928&
			4.6414004374753	&
			0.1865&
			0.361928840545e-3\\
			$q_1$ &
			78.0024790676476&
			101.559970448744&
			4.72003969924994&
			0.152&
			0.596292734846e-3\\
			$q_2$ &
			68.9691363164894&
			106.458487631437&
			4.76206349389489&
			0.0855&
			0.516631425667e-3\\
			$q_3$ &
			107.652359029517&
			131.731939948662&
			4.6869585791114&
			0.1325&
			0.439170010158e-3\\
			$q_4$ &
			69.6772856432126&
			42.6429340642319&
			4.92411292706019&
			0.072&
			0.455529757315e-3\\
			\hline
		\end{tabular}
	}
	\npnoround
	\label{tab:error_backends}
\end{table}

In Qiskit, backends represent either a simulator or a real quantum computer, and are responsible for running quantum circuits and returning the experimental results \cite{noauthor_ibm_nodate-1}. 
For each real quantum computer, a calibration datasheet is provided with up-to-date values\footnote{Available at \href{https://quantum-computing.ibm.com/}{https://quantum-computing.ibm.com}.} as the one presented in Table \ref{tab:error_backends}. 
Of the overall provided backends provided, we used the $1$-qubit ``Armonk'' and the $5$-qubit ``Burlington'' quantum computers. 
These backends can be accessed through a cloud-based queue, which can take several seconds or many minutes, depending on the previous jobs already in the queue. 
The main Qiskit simulation backend is the QASM Simulator, which emulates the execution of the quantum circuits on the local classical device. 
While it can be loaded with approximate noise models based on the calibration parameters of actual hardware devices, we decided not to introduce any noise in order to have a simulation reference baseline.

We run our circuits on the two quantum backends and on the QASM simulator, then we compare the obtained results. 
Since the maximum value of $N$ allowed by IBMQ in real backends is $N=2^{13}$, we maintained the value constant for each of the three sessions.

\subsection{Tests}

In order to obtain data for various simulation runs, we defined a sequence dataset $\mathbb{S}$ as
\begin{equation}
\mathbb{S}(s)
=
\left\lbrace
\left.
\Sigma 
\text{ s.t. }
\tau_{\ket{1}}
= 
\frac{s-1}{\tau}\;i
\;\;
\right\rvert
\;
i\in[0,s-1]
\right\rbrace,
\end{equation}
with $s$ being the number of samples in $[0, \tau]$. 
For example, for $\tau = 18$ we have
\begin{equation}
\mathbb{S}(3)
= 
\begin{bmatrix*}[l]
\Sigma \text{ s.t. } \tau_{\ket{1}}=\;0\\
\Sigma \text{ s.t. } \tau_{\ket{1}}=\;9\\
\Sigma \text{ s.t. } \tau_{\ket{1}}=18\\
\end{bmatrix*},
\end{equation}
in which every sequence $\Sigma$ is randomly generated to meet the $\tau_{\ket{1}}$ requirement. 
In order to provide a more consistent analysis, we repeated the $N$ experiments $n = 30$ times, having then $n$ iterations and $nN$ experiments for every sequence in the dataset. 
This lead to a total amount of $\tau n N$ experiments for each backend given a dataset $\mathbb{S}$.

\begin{figure}[p]
	\centering
	\scalebox{\plotscalingfactor}{
		\input{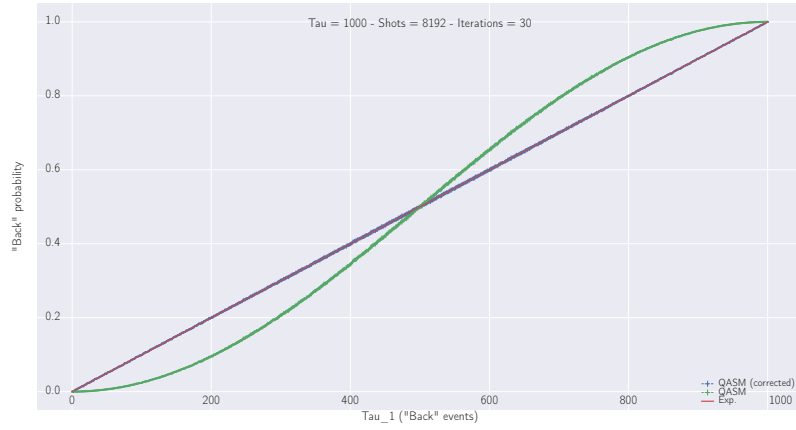}
	}	
	\caption{
		Results for $\tau=1000$, $\mathbb{S}(1000)$.
		Average values are plotted along with the corresponding standard deviations.
		QASM ``uncorrected'' results are the raw frequency outputs $\widehat{\;|c_1|^2}$.
		QASM ``corrected'' results are the empirical frequencies $\widehat{f}_{\ket{1}}$.
		Expected results are the expected relative frequency $f_{\ket{1}}$.
	}
	\label{fig:qubit1_dscr_states2_results_sim}
\end{figure}

We processed two datasets: 
$\mathbb{S}(1000)$ with $\tau=1000$, and $\mathbb{S}(10)$ with $\tau=10$. 
Due to the high amount of experiments needed for the first dataset, i.e., $\tau n N = 2.4576\cdot10^{8}$ the tests have been executed only offline via QASM simulations. 
Results provide information about the reliability of QASM simulations. 
In Fig. \ref{fig:qubit1_dscr_states2_results_sim} we compare the expected values with simulation results. 
The average values with the corresponding standard deviation are plotted. 
The ``uncorrected'' results are obtained plotting the raw frequency output of the experiments, namely $\widehat{\;|c_1|^2}$, without converting it with \eqref{eq:correction}. 
The ``corrected'' results are instead the ones obtained by plotting $\widehat{f}_{\ket{1}}$. 
The expected values are obtained by plotting the expected relative frequency of $f_{\ket{1}}$. 
These results are in line with theoretical expectations. 
The ``uncorrected'' results behavior has been reported as well because the non-linearity of these raw results may not be necessarily a bad thing for the case study we consider. 
In fact, this nonlinearity has a definite behavior that has been expressed in \eqref{eq:correction} and may represent a certain \emph{cognitive bias} introduced in the robot perceptual system.

\begin{table}[p]
	\centering
	\caption{
		Results for $\tau=10$, $\mathbb{S}(10)$.
		Average errors for each backend.}
	\npdecimalsign{.}
	\nprounddigits{3}
	\tabcolsep2mm 
	\renewcommand{\arraystretch}{1.05}
	\resizebox{\linewidth}{!}{%
		\begin{tabular}{ r 
				N{1}{3} 
				N{1}{3} 
				N{1}{3} 
				N{1}{3} 
				N{1}{3} 
				N{1}{3} 
				N{1}{3} 
				N{1}{3} 
				N{1}{3} 
				N{1}{3} 
				N{1}{3} }
			\hline
			$\bm{\tau_{\ket{1}}}$&
			\textbf{0} &
			\textbf{1} &
			\textbf{2} &
			\textbf{3} &
			\textbf{4} &
			\textbf{5} &
			\textbf{6} &
			\textbf{7} &
			\textbf{8} &
			\textbf{9} &
			\textbf{10} 
			\\
			\hline
			$\bm{\varepsilon_{qasm}}$& 
			{\fontseries{b}\selectfont}0 & 
			{\fontseries{b}\selectfont}0.12496694183849189 e-2& 
			{\fontseries{b}\selectfont}0.616708032641633  e-4& 
			{\fontseries{b}\selectfont}0.31053104866551884 e-3& 
			{\fontseries{b}\selectfont}0.36765593267351626 e-3& 
			{\fontseries{b}\selectfont}0.5957948750789921 e-4& 
			{\fontseries{b}\selectfont}0.8801358038016405 e-3& 
			{\fontseries{b}\selectfont}0.4226923865220389 e-3& 
			{\fontseries{b}\selectfont}0.6398747822171647 e-3& 
			{\fontseries{b}\selectfont}0.6827260459103535 e-3& 
			{\fontseries{b}\selectfont}0 
			\\ 
			{\footnotesize $f^{\ket{1}}-\hat{\;|c_1|^2}$}&
			0.0 & 
			0.7613118489583334 & 
			0.10456542968750002 & 
			0.09428710937499998 & 
			0.05395914713541666 & 
			0.9358723958335924 e-4& 
			0.05582275390625002 & 
			0.09442952473958333 & 
			0.10509847005208328 & 
			0.07585856119791667 & 
			0.0 
			\\
			\hline
			$\bm{\varepsilon_{burlington}}$& 
			{\fontseries{b}\selectfont}0.1639956877781075 & 
			{\fontseries{b}\selectfont}0.04063229650810926 & 
			{\fontseries{b}\selectfont}0.006287602573261913 & 
			{\fontseries{b}\selectfont}0.0015009119595953968 & 
			{\fontseries{b}\selectfont}0.007879508344412645 & 
			{\fontseries{b}\selectfont}0.01694671190813296 & 
			{\fontseries{b}\selectfont}0.02672188839355294 & 
			{\fontseries{b}\selectfont}0.03835079932963115 & 
			{\fontseries{b}\selectfont}0.07121355730546053 & 
			{\fontseries{b}\selectfont}0.08300412939874846 & 
			{\fontseries{b}\selectfont}0.13002405565762576 
			\\ 
			{\footnotesize $f^{\ket{1}}-\hat{\;|c_1|^2}$}& 
			0.06490478515625 & 
			0.051989746093750004 & 
			0.09862467447916667 & 
			0.09579671223958333 & 
			0.06623128255208338 & 
			0.026607259114583337 & 
			0.014090983072916674 & 
			0.043143717447916674 & 
			0.029223632812499956 & 
			0.019616699218749978 & 
			0.0411376953125 
			\\ 
			\hline
			$\bm{\varepsilon_{armonk}}$& 
			{\fontseries{b}\selectfont}0.13294089288128783 & 
			{\fontseries{b}\selectfont}0.0648663177110424 & 
			{\fontseries{b}\selectfont}0.028246114121911053 & 
			{\fontseries{b}\selectfont}0.014692836593116876 & 
			{\fontseries{b}\selectfont}0.0009365361972060726 & 
			{\fontseries{b}\selectfont}0.024672472470799733 & 
			{\fontseries{b}\selectfont}0.037309029354707035 & 
			{\fontseries{b}\selectfont}0.049166075958173905 & 
			{\fontseries{b}\selectfont}0.07556566973984902 & 
			{\fontseries{b}\selectfont}0.12443576015555524 & 
			{\fontseries{b}\selectfont}0.19230886625244992 
			\\
			{\footnotesize $f^{\ket{1}}-\hat{\;|c_1|^2}$}& 
			0.04297688802083333 & 
			0.03441975911458334 & 
			0.07687174479166668 & 
			0.07491455078124998 & 
			0.05310872395833338 & 
			0.038716634114583315 & 
			0.002160644531249978 & 
			0.02816162109375009 & 
			0.0240478515625 & 
			0.019222005208333393 & 
			0.08850911458333333 
			\\ 
			\hline
		\end{tabular}
	}
	\npnoround
	\label{tab:error}
\end{table}

\begin{figure}[p]
	\centering
	\scalebox{\plotscalingfactor}{\input{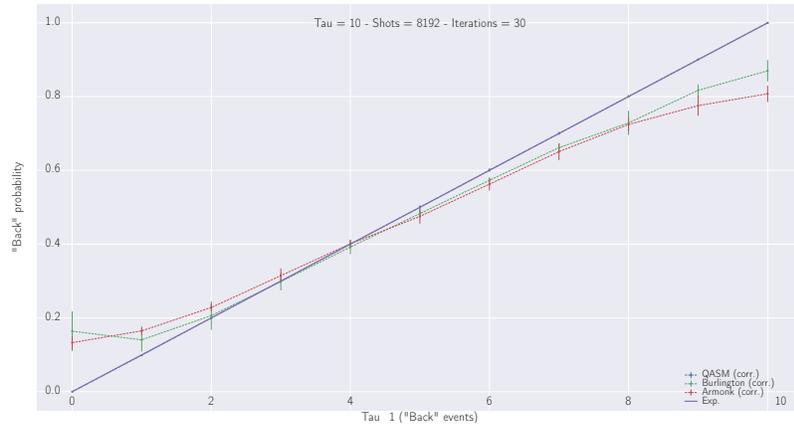}}
	\caption{
		Results for $\tau=10$, $\mathbb{S}(10)$.
		Average values are plotted along with the corresponding standard deviations.
		``Corrected'' results are the empirical frequencies $\widehat{f}_{\ket{1}}$ for each backend.
		Expected results are the expected relative frequency $f_{\ket{1}}$.
	}
	\label{fig:qubit1_dscr_states2_results_meas_corr}
\end{figure}

The tests for $\tau=10$, $\mathbb{S}(10)$ have been executed offline via QASM simulations as well as online via Armonk and Burlington backends. 
The average values with the corresponding standard deviation are reported in Fig. \ref{fig:qubit1_dscr_states2_results_meas_corr}. 
The corresponding errors are reported in Table \ref{tab:error}, which reports as well the error computed with respect to raw results $\widehat{|c_i|^2}$. 
Also in this case, results generated with QASM are in line with theoretical expectations. 
Regarding the results obtained by Armonk and Burlington backends, there are significant differences between the expectations and the measurements, differences which increase closer to the bounds $f_{\ket{1}}=0$ and $f_{\ket{1}}=1$, where there the ratio $\tau_{\ket{1}}:\tau_{\ket{0}}$ is not balanced anymore.


\section{Discussion and Workplan}
\label{sec:discussion}

\begin{figure}[t]
	\centering
	\includegraphics[width=8cm]{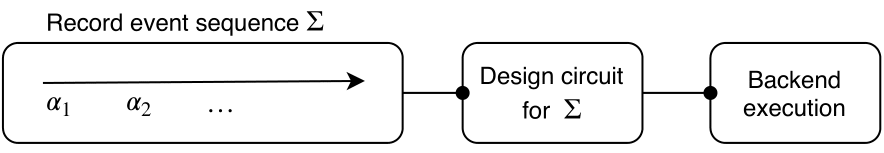}
	\caption{Workflow for sequence classification.}
	\label{fig:workflowcircuit}
\end{figure}

\begin{figure}[t]
	\centering
	\includegraphics[width=8cm]{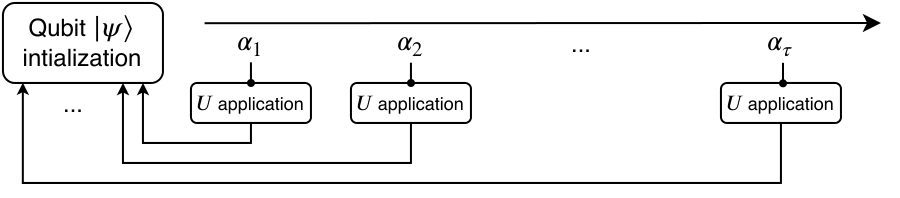}
	\caption{Workflow for online state update.}
	\label{fig:workflowrealtime}
\end{figure}

The obtained results seem to confirm the feasibility of our approach. 
Given the limitations of the considered case study, we nonetheless obtained the expected behavior through QASM simulations. 
With IBMQ backends we encountered errors for unbalanced ratio situations. 
This could be due to the fact that the IBMQ platform relies on circuit design and successive execution, and therefore it is not optimally calibrated for applications like the one presented here. 
The IBMQ architecture allows us to test the model, but only for \textit{a posteriori} sequences. 
In fact, after each sequence, a circuit was simulated and the results collected. 
If we were to make an \textit{a posteriori} classification task this could be enough. 
However, for online robot behavior, we need a framework that allows operator definition and synchronous application to a qubit, which represents the current state of the robot. 
Instead of a sequence classification workflow as the one illustrated in Fig. \ref{fig:workflowcircuit}, further research will have to investigate online state update applications implementing a workflow like the one illustrated in Fig. \ref{fig:workflowrealtime}.

It is preferable, in our opinion, to proceed with simulation-oriented techniques rather than actual realizations on quantum backends for two reasons. 
The first, as this preliminary study highlights, is that these services are not optimized for such unprecedented applications. 
The second is related to the fact that the exclusivity of these devices allows us to exploit them only by queue-based cloud services like IBMQ. 
This does not allow us to introduce it into a responsive runtime simulation. 

Our suggestion is to start to analyze information exploitation and interpolation for $n$-qubit multi-sensory systems, relying on simulations rather than physical realizations, to study which advantages a quantum-like approach can actually provide in robots' knowledge representation and reasoning.

%


\bibliographystyle{IEEEtran}
\bibliography{0.myzoterolib.bib}

\end{document}